%% file: ICRA_2026.tex
\documentclass[letterpaper, 10 pt, conference]{ieeeconf} 

\IEEEoverridecommandlockouts                              

\usepackage{amsmath} 
\usepackage{amssymb}  
\usepackage{graphicx}
\usepackage{balance}
\usepackage{lipsum}
\usepackage{algpseudocode}
\usepackage{multirow}
\usepackage{dsfont}
\usepackage{booktabs}
\usepackage{xcolor}
\usepackage{adjustbox}
\usepackage{stmaryrd}
\usepackage{graphicx} 
\usepackage{multirow}
\usepackage{nicefrac, xfrac}
\usepackage{amsmath} 
\usepackage{amssymb}  
\usepackage{xcolor}
\usepackage{lipsum}
\usepackage{booktabs}
\usepackage{hyperref}
\usepackage{cleveref}
\usepackage{adjustbox}
\usepackage{mathtools}
\usepackage{float}
\usepackage{amssymb}
\usepackage{pifont}
\usepackage{booktabs,adjustbox}
\usepackage{siunitx}
\usepackage{arydshln}

\usepackage{booktabs}
\usepackage{adjustbox}
\usepackage{xcolor}
\usepackage{graphicx}
\usepackage{amsmath}
\usepackage{float}
\usepackage{caption}
\usepackage{lipsum}
\usepackage{array} 
\usepackage{subcaption}
\usepackage{balance}
\usepackage{siunitx}

\usepackage{amsmath}

\usepackage[ruled,vlined,linesnumbered]{algorithm2e}
\definecolor{ForestGreen}{rgb}{0.133, 0.545, 0.133}

\SetCommentSty{mycommfont}

\definecolor{feicolor}{RGB}{201, 104, 104}

\title{\LARGE
Towards Autonomous Tape Handling for Robotic Wound Redressing
}
\author{Xiao Liang$^1$, Lu Shen$^2$, Peihan Zhang$^1$, Soofiyan Atar$^1$, Florian Richter$^1$, Michael Yip$^1$
\thanks{$^{1}$ Department of Electrical and Computer Engineering, University of California San Diego, La Jolla, CA 92093, USA {\tt\small \{x5liang, pez004, satar, frichter, yip\}@ucsd.edu}}
\thanks{$^{2}$ Sibley School of Mechanical and Aerospace Engineering, Cornell University, Ithaca, NY 14853, USA {\tt\small ls2244@cornell.edu}}}

\begin{document}
\newcommand{\te}{e}
\newcommand{\rotaxis}{\mathbf{v}}
\newcommand{\argmin}{\mathop{\mathrm{argmin}}} 
\newcommand{\argmax}{\mathop{\mathrm{argmax}}} 
\newcommand{\p}{\mathbf{p}}

\renewcommand*{\figureautorefname}{Fig.}
\renewcommand*{\subsectionautorefname}{Sec}
\renewcommand*{\sectionautorefname}{Sec}
\renewcommand*{\equationautorefname}{Eqn.}
\renewcommand*{\tableautorefname}{Table.}
\renewcommand*{\algorithmautorefname}{Alg.}

\newcommand{\uparrowgreen}{\textcolor{ForestGreen}{\textuparrow}}
\newcommand{\downarrowgreen}{\textcolor{ForestGreen}{\textdownarrow}}

\maketitle
\thispagestyle{empty}
\pagestyle{empty}

\begin{abstract}
Chronic wounds, such as diabetic, pressure, and venous ulcers, affect over 6.5 million patients in the United States alone and generate an annual cost exceeding \$25 billion. Despite this burden, chronic wound care remains a routine yet manual process performed exclusively by trained clinicians due to its critical safety demands. We envision a future in which robotics and automation support wound care to lower costs and enhance patient outcomes. This paper introduces an autonomous framework for one of the most fundamental yet challenging subtasks in wound redressing: adhesive tape manipulation. Specifically, we address two critical capabilities: tape initial detachment (TID) and secure tape placement. To handle the complex adhesive dynamics of detachment, we propose a force-feedback imitation learning approach trained from human teleoperation demonstrations. For tape placement, we develop a numerical trajectory optimization method based to ensure smooth adhesion and wrinkle-free application across diverse anatomical surfaces. We validate these methods through extensive experiments, demonstrating reliable performance in both quantitative evaluations and integrated wound redressing pipelines. Our results establish tape manipulation as an essential step toward practical robotic wound care automation.
\end{abstract}

\input{icra_tex_files/introduction}

\input{icra_tex_files/related_work}

\input{icra_tex_files/method}

\input{icra_tex_files/experiments}
\input{icra_tex_files/conclusion}

\balance
\bibliographystyle{IEEEtran}
\bibliography{root}

\end{document}

%% file: icra_tex_files/introduction.tex
\section{INTRODUCTION}
Chronic wounds, including diabetic ulcers, pressure ulcers, and ulcers secondary to venous hypertension, affects more than 6.5 million patients \cite{olsson2019humanistic} and a yearly cost of more than \$25 billion in the United States alone \cite{woundcare, han2017chronic}.
This growing demand for wound care coincides with a critical shortage of professional nursing staff, projected to reach over one million \cite{ghafoor2021impact}.
The majority of wound care occurs at a patient's home
\cite{lindholm2016wound}, which is especially crucial because most of the healing (and most complications) happen between clinic visits. However, home-based wound care not only strains the professional workforce due to prolonged travel and scheduling demands, but also places a substantial burden on family members who frequently assume caregiving responsibilities, as these informal caregivers must often sacrifice personal time and well-being to provide essential care, leading to emotional and physical stress \cite{chen2025family}. 

Automation technologies have been increasingly applied to address this global healthcare challenge from multiple perspectives. For example, immersive display systems have been explored for pain management during wound care \cite{maani2011virtual}. Advances in computer vision and machine learning have enabled automated wound assessment and healing prediction \cite{chen2024progress}. In robotics, researchers have investigated systems for autonomous wound care, including robotic debridement \cite{debridement} and automated bandaging \cite{bandaging}. Complementary automation approaches have also targeted wound tracking \cite{simango2023robotic} and three-dimensional reconstruction of wound geometry \cite{filko2025autonomous}.

More recently, \cite{liang2024autopeel} demonstrates the feasibility of robotic wound care in broader clinical and home settings by introducing an autonomous planning framework for safe wound dressing removal using general-purpose robot manipulators.
This approach is a strong start to the challenge problem of wound dressing, but ultimately only a single step of the overall wound redressing task, removing the initial adhesive tape given an initial grasp point on the tape.
In practice, wound care involves (1) initiating detachment and grasping of the adhesive tape, (2) removing the secondary dressing, (3) cleaning the wound and reapplying a new primary dressing, and (4) replacing the secondary dressing and securing it with medical tape. 
A common thread across these steps is the essential skill of adhesive tape manipulation. This is a critical capability where robots must avoid pitfalls such as unintended self-adhesion or excessive tension on fragile tissue.
For example, even the initial act of detaching tape, which is assumed in \cite{liang2024autopeel}, illustrates the complexity of this challenge: create a small detachment by scratching or prying at the edge if the tape and carefully enlarging the separation to achieve a precise and stable grasp. 

\begin{figure}[t]
    \centering
\includegraphics[width=0.48\textwidth]{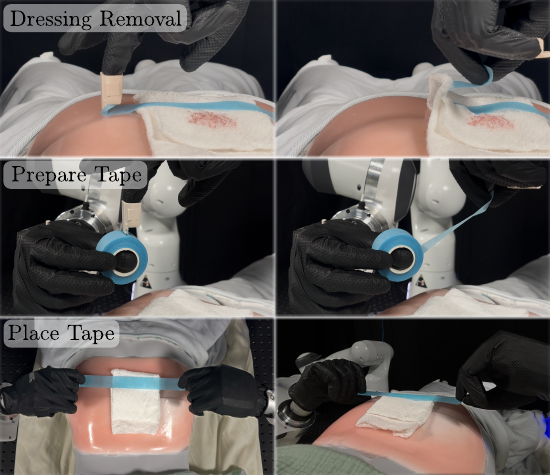} 
    \caption{This work enables advanced tape manipulation skills such as tape initial detachment (TID), and tape placement. They help automating three critical steps for wound redressing, including dressing removal, tape preparation, and tape placement to secure secondary dressings.}
    \label{fig:cover_photo}
    \vspace{-1.5em}
\end{figure}
\begin{figure*}[t]
    \centering
    \includegraphics[width=\textwidth]{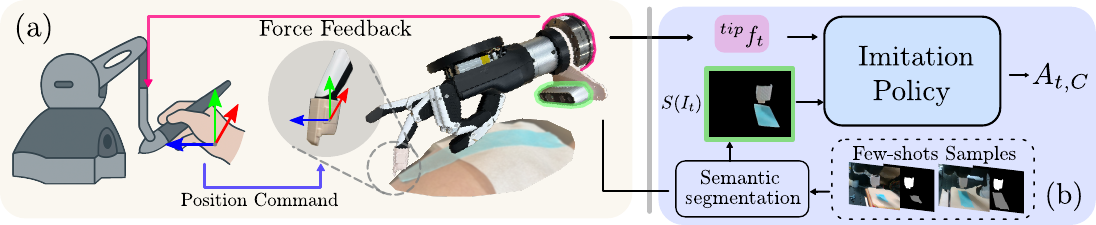}
    \vspace{-1.5em}
    \caption{The tape initial detachment framework includes (a) demonstration data collection via human teleoperation and (b) imitation learning for training and inference. An operator uses a haptic device to control the robot tip, receiving force feedback from a sensor between the hand and arm, while in-hand cameras capture images. The network is trained on segmented image and force data to predict future tip movements or forces, with semantic segmentation initialized from a small set of labeled samples.}
    \vspace{-1.5em}
\label{fig:teleop_and_learning}
\end{figure*}

In this paper, we present the first autonomous approach for medical tape manipulation designed to enable wound redressing, which directly addresses the challenges of initiating detachment (part of step 1) and re-securing dressings with new tape (step 4) --- further progressing the field towards fully end-to-end autonomous wound care dressing.
We use two separate approaches towards solving tape manipulation: a learning based approach for tape initial detachment (TID) and trajectory optimization using a differentiable simulation for precise tape placement.
This hybrid design reflects the nature of the task: while the detachment phase involves complex, hard-to-model adhesive dynamics best addressed with learning, the placement phase requires accurate sensing and careful planning with respect to different body parts, where traditional control offers greater generalization and reliability. Our contributions are summarized as follow:
\begin{itemize}
    \item A force-feedback-enabled teleoperation platform for collecting demonstrations of TID, coupled with a visual imitation learning that leverages semantic segmentation to isolate task-relevant information to improve generalization.  
    \item A numerical trajectory optimization method for achieving optimal robot trajectories and force profiles for secure medical tape placement while considering varying surface skin geometries.
    \item Extensive experiments, including quantitative evaluations and a full pipeline demonstration, validating robotic tape manipulation as a key capability for enabling practical wound redressing automation.
\end{itemize}

%% file: icra_tex_files/related_work.tex
\section{Related Works}

\subsection{Deformable and Adhesive Materials Manipulation}
Robotic manipulation of deformable objects has been an active research area due to its importance in tasks such as cloth manipulation \cite{hietala2022learning}, bagging \cite{10341841}, and tissue manipulation \cite{liang2024real, liang2024medic} for robotic surgeries. Similarly, manipulation of linear objects such as ropes and cables has been explored with techniques ranging from physics-based modeling \cite{10093017} to learning-based policies \cite{10610754}. These studies highlight the challenges of reasoning about high-dimensional configuration spaces and contact-rich interactions. Beyond general deformables, researchers have also begun to address robot manipulation of adhesive materials, where the dynamics are further complicated by adhesion forces and nonlinear detachment mechanics. For example, peeling and stripping of Velcro was studied in \cite{9561030}. In the medical domain, automatic peeling of adhesive tapes has been investigated for wound care applications \cite{liang2024autopeel}. Active manipulation of deformables to reveal potential adhesion conditions was explored in \cite{shinde2024jiggle}. Compared to cloth and rope, adhesive manipulation introduces unique challenges such as controlling peel angle, velocity, and force to safely manage detachment, which are especially critical in medical contexts.

\subsection{Robotic Systems for Home Healthcare and Assistive Care} 
Robotics has also been explored as a means to support healthcare delivery in home and assistive settings. A large body of work has focused on assistive robots for activities of daily living, including feeding \cite{candeias2018vision}, dressing \cite{10465608}, and mobility assistance \cite{xing2021admittance}. In parallel, telepresence and socially assistive robots have been developed to facilitate remote monitoring, companionship, and emotional support for patients in the home \cite{9223470, dickey2025advances}. Humanoid platforms have recently become an enabling platform technology that allows for exploring more diverse applications that would otherwise not be reasonable to have a specialized robot perform. More recently, \cite{atar2025humanoids} demonstrated that humanoid robots and their human-like hands can manipulate various tools and objects, performing medical tasks such as auscultation and ultrasonography, demonstrating both their generality as well as their potential to transition from clinical environments into home care. Despite these advances, wound care remains underexplored, despite being one of the most labor-intensive and sensitive aspects of caregiving.

%% file: icra_tex_files/method.tex
\section{METHODS}
\subsection{Learning for Tape Initial Detachment (TID)}

Detaching tape when it is still adhered to the skin or dressing constitutes the fundamental precondition for all subsequent tape manipulation tasks. This problem involves handling a deformable, thin, and adhesive medium under complex physical interactions. The robot must apply localized peeling forces at the tape edge to initiate detachment, while carefully regulating the end-effector’s velocity to control the propagation of the peeling front. 
Knowing the limitations of simulation and model-based approaches, we adopt a data-driven approach to model the tape–skin interaction. Specifically, an imitation learning framework is used to acquire interaction strategies from a dataset of human teleoperation demonstrations. 

Our teleoperation system for data collection to conduct imitation learning framework is shown in \autoref{fig:teleop_and_learning}. 
Since most robotic hands do not include fingernails, the robotic dexterous hand used in this study (RH56DFTP, Inspire Robots) was augmented with a 3D-printed finger cap, allowing it to scratch along the tape edge similar to a human finger nail. The dexterous hand mounts to a robot manipulator arm along with a force/torque sensor (Axia80, ATI Technologies), and a human operator controls the position of the dexterous hand's finger tip through interfacing with a haptic teleoperation device (Phantom Omni, Sensable Technologies), while receiving force feedback from the force/torque sensor. For visual data, an RGB camera (Intel RealSense) is mounted under the wrist of the dexterous hand.

Using the physical setup above, a human teleoperation dataset $D:\{\tau_i, ...,\tau_N\}$ is collected, where each demonstration $\tau_i = \{(o_0, x_0, a_0), ... (o_T, x_T, a_T)\}$ is a collection of observation $o_t$, proprioception $s_t$, and actions $a_t$. 
The observation, $o_t :\{^{tip}f_t, \mathbf{S}(I_{t
})\}$, contains the force measurement at the finger tip, $^{tip}f_t \in \mathbb{R}^3$,  and segmentation of the finger tip and tape, $S(I_{t
})$, from the RGB image, $I_t$, from the in-hand camera. 
The segmentation is adopted to generalize other different backgrounds by keeping only task-relevant portion of the image.
Previous work \cite{cheng2023putting} is used to enable this module. It keeps a small few-shot dataset as memory and infer segmentation masks for new images.
$x_t$ includes the force reading from the force/torque sensor in between the dexterous hand and the robot manipulator.
The objective of imitation learning is to learn policy $\pi: o_t \mapsto A_{t, C}$, where $A_{t, C}$ represents a future sequence of tip frame-centric position actions
\begin{equation}
    A_{t,C} = \left\{ \Delta p_s^{\text{tip}} \;\middle|\; s \in [t, t+C] \right\} \in \mathbb{R}^{3 \times C}.
\end{equation}
Each action is $\Delta p_t^\text{tip} = p^{\text{tip}}_{t+1} - p^{\text{tip}}_{t}$.
Finally, an Action Chunking Transformer (ACT) \cite{zhao2023learning} architecture is used to train the imitation learning policy.

\begin{figure}[t]
    \centering
    \includegraphics[width=0.5\textwidth]{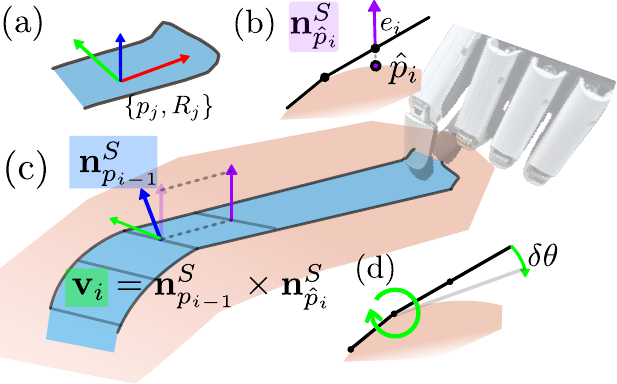} \vspace{-1.em}
    \caption{Key geometric concepts that the proposed tape placing method considers. (a) illustrates a tape element's position and orientation.
    In each numerical step, (b) the algorithm finds the next free tape element's closest point $\hat p_i$ and its surface normal $n^S_{\hat p_i}$. (c) Then it computes an rotation axis $\rotaxis_i$ with cross product between the normal vector at the latest attachment element $n^S_{p_{i-1}}$ and $n^S_{\hat p_i}$. (d) In next step, the tape region is rotated around $\rotaxis_i$ to approach the surface.}
    \label{fig:tpae_placing_geometry}
    \vspace{-1.em}
\end{figure}

\begin{algorithm}[!t]
\small
    \caption{Numerical Solution for Bimanual Tape Placing Pose}
    \label{alg:numerical_pose}
    
    \SetKwInOut{Input}{Input}
    \SetKwInOut{Output}{Output}
    
    \Input{Skin surface $S$, Initial adhesion point $p_{\text{init}}$, Initial tape direction $\mathbf{d}_{\text{init}}$, tape length $l$, element length $l_e$, angle step $\delta \theta$, adhesion threshold $\varepsilon$}
    \Output{Start-side pose trajectory $\mathcal{P}_s$, End-side pose trajectory $\mathcal{P}_e$, Start-side tension direction $\mathcal{F}_s$, End-side tension direction $\mathcal{F}_e$}

    \BlankLine

    $N \gets \text{round}(l / l_e)$\;
    $i_{\text{mid}} \gets \lfloor N / 2 \rfloor$\;
    \For{$j \in \{0, \dots, N-1\}$}{
        $p_j^{\text{init}} \gets p_{\text{init}} + (j - i_{\text{mid}}) \cdot l_e \cdot \mathbf{d}_{\text{init}}$\;
    }
    $p \gets p^{\text{init}}$\;
    $p_{i_{\text{mid}}} \gets \operatorname{ClosestPointOnSurface}(S, p_{i_{\text{mid}}})$\;
    
    \BlankLine
    
    $i_s, i_e \gets i_{\text{mid}}, i_{\text{mid}}$\;
    $R_{s, \text{acc}}, R_{e, \text{acc}} \gets \mathbf{I}_3, \mathbf{I}_3$\;
    $\mathbf{v}_s, \mathbf{v}_e \gets \textit{InitialAxis}(S, p_{i_{\text{mid}}}, \mathbf{d}_{\text{init}})$\;
    $\mathcal{P}_s, \mathcal{P}_e, \mathcal{F}_s, \mathcal{F}_e \gets \{\}, \{\}, \{\}, \{\}$\;
    
    \BlankLine
    
    \While{$i_s > 0$ \textbf{or} $i_e < N-1$}{
        \If{$i_e < N-1$}{
            $p_{\text{pivot}} \gets p_{i_e}$\;
            \For{$j \in \{i_e+1, \dots, N-1\}$}{
                $p_j \gets p_{\text{pivot}} + R_{e, \text{acc}} \cdot (p_j^{\text{init}} - p_{i_e}^{\text{init}})$\;
            }
            
            \If{\text{distance}$_S(p_{i_e+1}) < \varepsilon$}{
                $i_e \gets i_e + 1$\;
                $p_{i_e} \gets \textit{ClosestPointOnSurface}(S, p_{i_e})$\;
            }
            \Else{
                \If{$i_e > i_{\text{mid}}$}{
                    $\mathbf{n}_{\text{curr}} \gets \textit{SurfaceNormal}(S, p_{i_e})$\;
                    $\mathbf{n}_{\text{prev}} \gets \textit{SurfaceNormal}(S, p_{i_e-1})$\;
                    $\mathbf{v}_{\text{new}} \gets \mathbf{n}_{\text{prev}} \times \mathbf{n}_{\text{curr}}$\;
                    \If{$\lVert \mathbf{v}_{\text{new}} \rVert \neq 0$}{ $\mathbf{v}_{e} \gets \mathbf{v}_{\text{new}} / \lVert \mathbf{v}_{\text{new}} \rVert$\; }
                }
                $\Delta R \gets \text{Rodrigues}(\mathbf{v}_{e}, \delta\theta)$\;
                $R_{e, \text{acc}} \gets \Delta R \cdot R_{e, \text{acc}}$\;
            }
        }
        
        \If{$i_s > 0$}{
            $p_{\text{pivot}} \gets p_{i_s}$\;
            $\cdots$\tcp{Symmetric logic for the start side (mirror line 13 $\sim$ 27)}
            
        }
        
        \BlankLine
        $\mathcal{P}_s \gets \mathcal{P}_s \cup \{ (p_0, R_{s, \text{acc}}) \}$\;
        $\mathcal{P}_e \gets \mathcal{P}_e \cup \{ (p_{N-1}, R_{e, \text{acc}}) \}$\;
        $\mathcal{F}_s \gets \mathcal{F}_s \cup \{ (p_0 - p_{i_s}) / \lVert p_0 - p_{i_s} \rVert \}$\;
        $\mathcal{F}_e \gets \mathcal{F}_e \cup \{ (p_{N-1} - p_{i_e}) / \lVert p_{N-1} - p_{i_e} \rVert \}$\;

    }
    
    \BlankLine
    \Return $\mathcal{P}_s, \mathcal{P}_e,\mathcal{F}_s,\mathcal{F}_e$
\end{algorithm}

\subsection{Trajectory Optimization for Tape Placement}

Medical tape is widely used in wound care to secure secondary dressings in place. For successful placement, the tape must adhere smoothly to the patient’s skin, conforming to varying geometries without forming wrinkles or sticking to itself. Achieving this requires precise motion planning and control of the tape’s interaction with both the underlying dressing and the skin surface, while maintaining appropriate tension to ensure reliable adhesion. 

A model of tape-tensioning and surface application is significantly less complex than the dynamics present in TID, and model-based methods are preferred if possible due to explainability. Thus, a numerical optimization approach based on a model definition for tape (visualized in \autoref{fig:tpae_placing_geometry}) is developed. In this case, the numerical method takes advantage of the model to simultaneously compute dense optimal robot end-effector pose and tension trajectories that lead to proper tape-to-surface adherence with no wrinkles. 
Let a tape segment with known length $l$ be represented as a set of discrete elements $E = \{\te_j\}^{J}_{j=0}$, where each element be defined by a position and orientation $\te_j = \{p_{j}, R_{j}\}$, and with $\te_i$ being the closest free tape element next to an attached element $\te_{i-1}$. The element $\te_{i:J}$ is simulated as a rigid body with planar and rectangular shape, and it rotates the $\te_{i:J}$ around a rotation axis $\mathbf{\hat v}_{i-1}$ passing through $p_{\te_{i-1}}$.  

The core idea of this method is that, at each numerical step, the method looks for the best robot action that brings the unattached tape portion $\te_{i:J}$ closer to the skin surface. At the same time, the chosen action must respect task-specific constraints such as tape inextensibility, where the element length of the tape stays constant: 
\begin{equation}
    C_{\text{length}}(E) = \te_j - \te_{j-1} - \frac{l}{J}.
\end{equation}
Another constraint restricts the $\te_{i:J}$ to be held in tensioned without slack:
\begin{equation}
    C_{\text{tension}}(\te_{i:J}) = \lVert p_J - p_{i-1} \rVert \;-\; \sum_{k=i}^{J} \lVert p_k - p_{k-1} \rVert  .
\end{equation}
Lastly, tape wrinkles need to be avoid by limiting rotational movement of $\te_{i:J}$ around the yaw-axis of tape element $\te_{i-1}$:
\begin{equation}
    C_{\text{wrinkle}}(\te_{i}) = \mathbf{v}_{i-1}^\top \mathbf{z}_{i-1},\  \texttt{where } \mathbf{z}_{i-1} = \mathbf{n}_{p_{i-1}}^S
\end{equation}
This constraint is satisfied by finding the correct rotation axis $\rotaxis_{i-1}$ that is orthogonal to the normal vector at $\te_{i-1}$. The process of computing $\rotaxis_{i-1}$ starts with finding the closest point $\hat p_{i}$ of $p_i$ on skin surface $S$, followed by computing the surface normal $\mathbf{n}^S_{\hat p_i}$. The correct rotation axis is computed as the following cross product,
\begin{equation}
    \begin{split}
\rotaxis_{i-1} =
\begin{cases}
\mathbf{n}^S_{p_{i-1}} \times \mathbf{n}^S_{\hat p_{i}}, & 
\text{if } \mathbf{n}^S_{p_{i-1}} \times \mathbf{n}^S_{\hat p_{i}} \neq \mathbf{0}, \\[6pt]
\rotaxis_{i-2}, & \text{otherwise.}
\end{cases}
    \end{split}
\end{equation}
$\rotaxis_{i-1}$ is set to the prior rotation axis $\rotaxis_{i-2}$ if two normal vectors are parallel, which is possible when the local surface is flat.
$\te_{i:J}$ rotates around $\rotaxis_{i-1}$ for an angle step $\delta \theta > 0$, bringing it closer to the surface $S$. For a element $e_k \in \te_{i:J}$, its pose is updated as
\begin{equation}
\begin{split}
        R_{\text{acc}, t} & = \text{Rodrigues}(\rotaxis_i, \delta \theta) \cdot R_{\text{acc}, t-1}, \\
    R_{k, t} & = R_{\text{acc}, t}, \\ 
    p_{k, t} & = p_{i-1} + R_{k, t} (p_{k, t} - p_{i-1}),
\end{split}
\end{equation}
Here, $R_{\text{acc},t}$ is an orientation accumulated over the numerical process, and $R_{\text{acc}, 0} = \mathbf{I}$.
At each numerical step, adhesion of $\te_i$ is determined by checking its collision the surface $S$ under the following condition:
\begin{equation}
    \texttt{Distance}_S(p_i) < \varepsilon,\ \varepsilon>0,
\end{equation}
where $\varepsilon$ is a small threshold.
If a collision is found, the $\te_i$ is viewed as adhered and will stay static later on. The numerical method iterates until the last tape element $\te_{J-1}$ is attached to the skin surface. In this process, the robot end effector pose can be obtained from the pose of $\te_{J-1, t}$, and the tension direction is $(\te_{J-1, t} - \te_{i-1}) / \|\te_{J-1, t} - \te_{i-1}\|$.

The above numerical method can be applied to bi-manual tape placing. Assume a desired tape position $p$ and direction $d$ are given, where $d$ is constrained to lie in the local tangent plane of the mesh, i.e., ${\mathbf{n}_p^{S}}^\top d = 0, \text{with } \mathbf{n}_p^S \text{ being the surface normal at } p.$
 Bi-manual tape placing is achieved by first placing the middle element of the tape at $p$ and aligning its initial direction to the desired taping direction $d$, while using symmetrical logic with the proposed numerical method to compute pose and tension trajectories for both ends independently as shown in Algorithm \autoref{alg:numerical_pose}. 
 
The tape length $l$ is measured by lightly tensioning it with two end-effectors pulling in opposite directions, then computing the distance between them.
During execution of the planned trajectories, an impedance control scheme is used to hold the tape taut by adding a small residual distance $d$ to the translation component of end-effector poses in the tension direction. Our experiments have shown the advantages of the impedance control scheme for compensating for discretization and motion errors, leading to better adherence.
Skin surface $S$ is represented by a triangular mesh reconstructed from an RGBD camera, and is post-processed to be convex. Note that our method can be extended to work with concave geometry by performing collision checking for all $\te_{i:J}$, and skipping to the last element in contact.


\subsection{Implementation Details}
The TID imitation policy was carried out on the RH56DFTP Inspire Robotic hands with the attached fingernail, mounted on Franka Emika Panda manipulators with a wrist-mounted Intel RealSense D435 camera streaming $640 \times 480$ RGB images. Each variant of the imitation learning method was trained with a dataset of 100 demonstrations collected via teleoperation combined with force-guided demonstrations. At each timestep, the system logged the RGB image, hand pose, hand force/torque, and the expert action (displacement and rotation increments). TID was performed by the robot hand with a 3D-printed nail on its middle finger: the fingertips established a small measured pre-tension and executed short, force-regulated detachment vectors to initiate peeling. The Axia80 force/torque sensor monitored contact throughout the process. Tape-placement experiments employed two 7-DoF Franka Emika Panda manipulators with their standard grippers. 
Two Azure Kinect units, mounted at complementary overhead and side angles, provided visual feedback. All cameras and robot bases were calibrated into a common world frame. All experiments used standard medical adhesive tape (3M Micropore, 25 mm width) cut to standardized lengths of 150–400 mm, depending on the application scenario. Gauze pads ($50 \times 50$ mm) were pre-positioned to simulate wound-dressing coverage requirements.

%% file: icra_tex_files/experiments.tex
\section{Experiments \& Results}


The proposed framework is evaluated across both tape initial detachment (TID) \autoref{sec:tid_eval} and tape placement \autoref{sec:tape_placing_eval} tasks in scenarios ranging from phantom surfaces and human skin to anatomical leg, hip, and heel models. Finally, we demonstrate the full wound redressing pipeline, combining TID and placement into a multi-step automation \autoref{sec:pipeline_demo}.
\begin{figure}[t]
    \centering
    \includegraphics[width=0.49\textwidth]{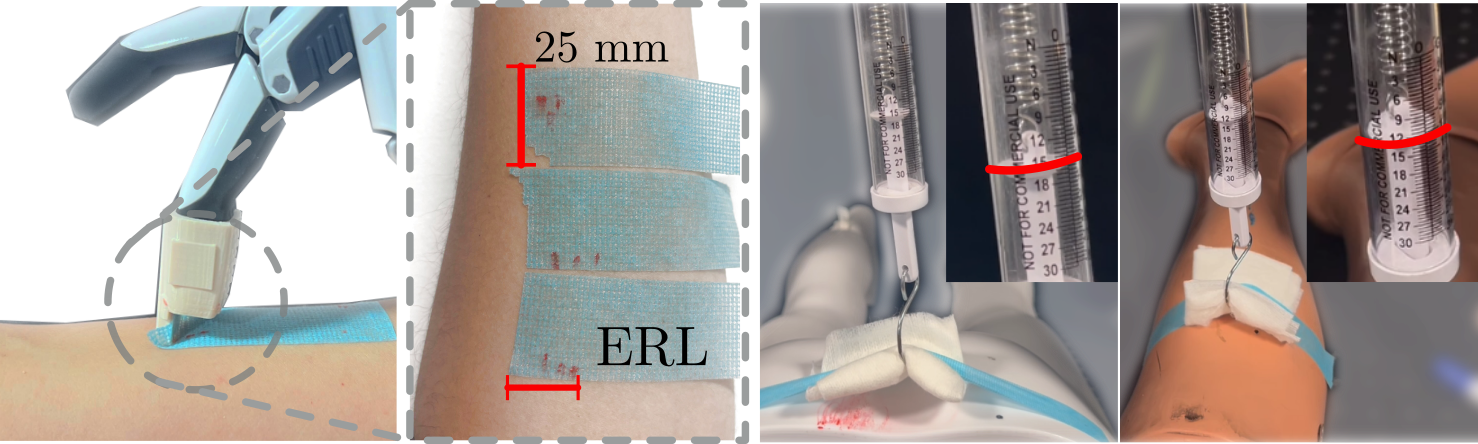} 
    \caption{Visualization of the quantitative evaluation metrics used in this work. Left: effective removal length (ERL) is measured by marking the furthest detached tape position from the tape end. Right: tape placement security is quantified by recording the maximum reading that a force gauge (N) when forcefully removing the tape and gauze.}
    \label{fig:experiment_metric}
    \vspace{-1em}
\end{figure}
\subsection{Results of Tape Initial Detachment}\label{sec:tid_eval}
TID performance is evaluated in two common wound care scenarios: (1) lifting the tape from human skin for dressing removal, in which a robot needs to be capable of making safe contact with diverse soft human skin, and (2) lifting it from a tape roll, which, in contrast, is a more rigid and curved surface. We train independent policies for different scenarios, as their material and adhesion properties are distinct. Demonstration datasets are collected on skin phantom, and tape roll of different radius.

\begin{table*}[t]
\centering
\caption{\textbf{Performance of different TID approaches across multiple scenarios.} Imitation learning (IL) achieved a higher effective removal length (ERL) and lower force profile than the motion planning (MP) baseline. IL methods also demonstrated the ability to perform well under out-of-distribution cases of human skins (arm). FF-IL, which is trained using a force-feedback-aware dataset achieved the equivalent performance to NF-IL that is trained without force-feedback dataset, while producing the least amount of forces that could harm the skin.}
\begin{adjustbox}{width=1.0\textwidth}
\scriptsize
\setlength\tabcolsep{6pt} 
\renewcommand{\arraystretch}{1.1}
\sisetup{
  table-number-alignment = center,
  detect-weight = true,
  detect-inline-weight = math
}
\begin{tabular}{l
  | c c
  | c c
  | c c
  | c c}
\toprule
& \multicolumn{2}{c|}{Phantom} 
& \multicolumn{2}{c|}{Human Skin} 
& \multicolumn{2}{c|}{Tape roll (20 mm)} 
& \multicolumn{2}{c}{Tape roll (25 mm)} \\
\cmidrule(lr){2-3} \cmidrule(lr){4-5} \cmidrule(lr){6-7} \cmidrule(lr){8-9}
& {ERL (mm) \uparrowgreen} & {Force (N) \downarrowgreen} 
& {ERL (mm) \uparrowgreen} & {Force (N) \downarrowgreen} 
& {ERL (mm) \uparrowgreen} & {Force (N) \downarrowgreen} 
& {ERL (mm) \uparrowgreen} & {Force (N) \downarrowgreen} \\
\midrule
MP 
& $9.8\pm6.0$ & $4.3\pm0.7$
&  $11.7\pm6.6$ & $2.6\pm0.7$
& - & $>10$
& - & $>10$\\
NF-IL 
& $\mathbf{19.5 \pm 7.1}$ & $5.2 \pm 1.4$ 
& $\mathbf{13.1 \pm 1.8}$  & $2.3 \pm 0.4$  
& $12.1\pm2.9$ & $5.0\pm1.0$ 
& $11.3\pm3.3$ & $5.9\pm1.0$ \\
FF-IL 
& $18.9 \pm 4.9$ & $\mathbf{3.1 \pm 0.4}$
& $12.0 \pm 2.2$  & $\mathbf{1.9\pm0.4}$  
& $\mathbf{12.4\pm1.9}$ & $\mathbf{3.3\pm0.8}$ 
& $\mathbf{11.8\pm1.9}$ & $\mathbf{3.7\pm0.6}$ \\
\bottomrule
\end{tabular}
\end{adjustbox}
\label{tbl:tid}
\end{table*}

The evaluation metrics for the TID policy were \textbf{effective removal length (ERL)}, measured in millimeters, which represents the length of the tape successfully detached and is shown in \autoref{fig:experiment_metric} and \textbf{force applied} externally to soft skin, measured by the force and torque sensor.
The proposed method was compared to a motion-planning-based (model-based) baseline method, denoted as \textit{MP}. This \textit{MP} method skeletonizes a binary mask of the tape to a polyline on the image frame, and then uses depth information to back-project the polyline into 3D space to compute a desired finger tip trajectory. The trajectory is then executed under position-based PID control.
An ablation study was performed to compare variants of the proposed methods, \textit{NF-IL} representing training without force feedback and \textit{FF-IL}or with force feedback. At the beginning of all experiments, the tip of the finger was moved to be roughly aligned with the tape, and each comparison method was repeated three times. Each execution was terminated when it exceeded a time limit of 10 seconds, then returning to the starting position.

\autoref{tbl:tid} compares the various methods, where every trials was repeated 10 times. On the phantom, NF-IL produced the most effective removal. FF-IL achieved a slightly shorter removal length but applied significantly lower force to the phantom skin. In comparison, the motion planning baseline led to a shorter removal length but a larger force. When evaluated in the out-of-distribution case on human arm skin, the same pattern still held. However, both imitation learning policies achieved less removal length, possibly due to changes in material and adhesion properties. Two tape rolls with radii of 20 mm and 25 mm were evaluated. In both scenarios, the baseline method failed due to controller failures when the contact force between the fingertip and the tape roll became too large. FF-IL outperformed NF-IL in both removal length and force. This was because NF-IL led to a greater frictional force during contact, making its control slow and ineffective. In contrast, FF-IL, trained with force-feedback-enabled data, was able to sweep smoothly through the roll surface, applying appropriate force to detach the tape.

In general, imitation learning–based policies outperformed the motion planning baseline in terms of both ERL and force metrics. In comparison, the policies trained with the force-feedback dataset were more desirable when considering both wound safety and removal effectiveness. 
The imitation learning methods achieved comparable performance on unseen cases involving human arms and tape rolls of various sizes, justifying the choice of using semantic segmentation to focus only on the task-relevant portions of the images. 

\begin{figure}[t]
    \centering
\includegraphics[width=0.49\textwidth]{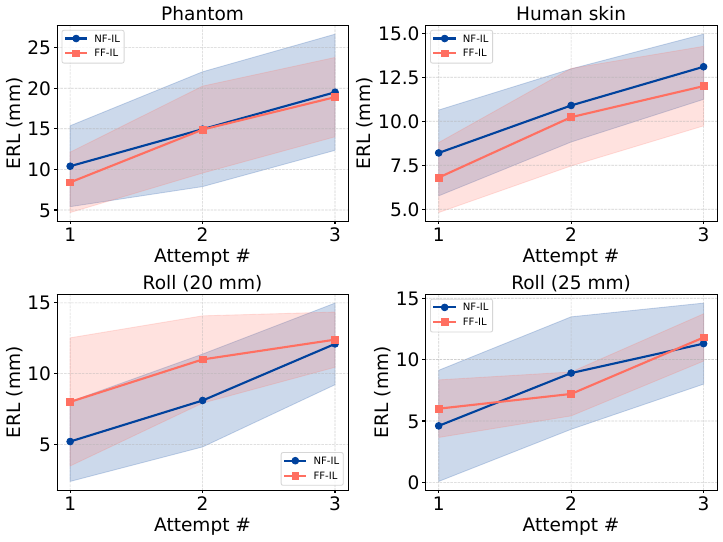} 
    \vspace{-1.5em}
    \caption{Relationship between the resulting ERL and number of attempts. Within 3 attempts, ERL of the proposed TID method linearly increases as the number of attempts increases in multiple scenarios, showing that its capability of improving outcome with repetition}
    \label{fig:erl_attempt}
    \vspace{-1.5em}
\end{figure}

An additional study investigated how multiple attempts led to more effective removal. \autoref{fig:erl_attempt} showed their relationship: within three attempts, the ERL of the proposed TID method increased almost linearly with the number of attempts across all testing scenarios. This highlighted the possibility of achieving further tape removal with additional attempts.

\begin{table*}[t]
\centering
\caption{\textbf{Tape placing effectiveness (first: coverage \%, second: maximum removal force).} The impedance control approach, particularly when incorporating residual physics, achieves significantly higher coverage than the baseline and simple position control policies. This approach also demonstrates the ability to generalize to challenging geometries like the sole and the heel, where other methods fail to achieve stable tape placement.}
\begin{adjustbox}{width=1.0\textwidth}
\scriptsize

\setlength\tabcolsep{6pt} 
\renewcommand{\arraystretch}{1.1}

\sisetup{
  table-number-alignment = center,
  detect-weight = true,
  detect-inline-weight = math
}

\begin{tabular}{l
  | c c
  | c c
  | c c
  | c c}
\toprule
& \multicolumn{2}{c|}{Leg} 
& \multicolumn{2}{c|}{Hip} 
& \multicolumn{2}{c|}{Sole} 
& \multicolumn{2}{c}{Heel} \\
\cmidrule(lr){2-3} \cmidrule(lr){4-5} \cmidrule(lr){6-7} \cmidrule(lr){8-9}
& {Coverage (\%) \uparrowgreen} & {MRF (N) \uparrowgreen} 
& {Coverage (\%) \uparrowgreen} & {MRF (N) \uparrowgreen} 
& {Coverage (\%) \uparrowgreen} & {MRF (N) \uparrowgreen} 
& {Coverage (\%) \uparrowgreen} & {MRF (N) \uparrowgreen} \\
\midrule
Baseline 
& 68.6 $\pm$ 5.2 & 8.3 $\pm$ 3.1
& 66.8 $\pm$ 6.8 & 17.3 $\pm$ 5.4
& 85.5 $\pm$ 4.9 & 4.3 $\pm$ 2.5
& 84.7 $\pm$ 4.6 & 0.6 $\pm$ 0.5 \\
Position 
& 82.3 $\pm$ 4.5 & 12.6 $\pm$ 4.2
& 83.7 $\pm$ 5.1 & 28.9 $\pm$ 8.9
& 88.3 $\pm$ 2.7 & 11.6 $\pm$ 2.7
& 87.9 $\pm$ 5.1 & 5.4 $\pm$ 2.3 \\
Impedance
& 75.8 $\pm$ 3.8 & 10.7 $\pm$ 2.5
& 81.0 $\pm$ 4.2 & 26.2 $\pm$ 6.1
& 84.7 $\pm$ 6.9 & 9.0 $\pm$ 3.5
& 89.3 $\pm$ 4.5 & 5.2 $\pm$ 2.2 \\
Impedance+residual
& \textbf{100 $\pm$ 0.0} & \textbf{23.0 $\pm$ 1.5}
& \textbf{95.4 $\pm$ 2.1} & {$>$\textbf{30} (--) }
& \textbf{89.9 $\pm$ 5.9} & \textbf{18.4 $\pm$ 2.8}
& \textbf{93.1 $\pm$ 4.4} & \textbf{10.2 $\pm$ 2.3} \\
\bottomrule
\end{tabular}

\end{adjustbox}
\label{tbl:tape_effectiveness}
\end{table*}

\subsection{Results of Tape Placing}\label{sec:tape_placing_eval}

Experiments for autonomous tape placement were conducted across multiple scenarios and surface geometries to evaluate the effectiveness of the approach in various real-world wound care applications.

\begin{figure}[t]
    \centering
    \includegraphics[width=0.48\textwidth]{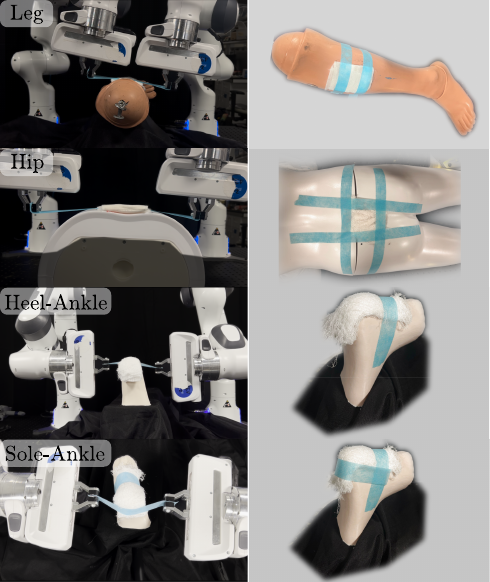}
    \caption{Results of the proposed tape placing method. 1st column: execution of one planned tape placing trajectory for each body geometry. 2nd column: outcome of multiple tapes placed consecutively by the proposed method. The number of tapes are 2, 4, 2 for leg, hip, and heel geometries.}
    \label{fig:tape_placing}
    \vspace{-1.5em}
\end{figure}

A robot workspace was set up to emulate realistic wound-care scenarios using anatomical surface models representing body regions commonly affected by chronic wounds and requiring dressing application. Three representative geometries with increasing complexity were considered, as shown in Fig.~\ref{fig:tape_placing}. The first was a curved cylindrical surface resembling the human leg, a typical site for \textit{venous ulcers}. The second was a hip geometry with moderate curvature and a larger surface area, which is frequently affected by \textit{pressure ulcers}. The third was a heel geometry, characterized by pronounced curvature and limited accessibility, representing a high-risk site for both \textit{pressure ulcers} and \textit{arterial ulcers}.


Due to the complex anatomy of the heel region and the variety of potential wound locations, we implemented and validated two distinct clinical scenarios. Specifically, we focused on the plantar sole and the posterior heel/Achilles region, as these sites are clinically recognized as highly vulnerable: plantar wounds are prone to high pressure and shear during weight bearing  \cite{NPIAP2019}, while posterior heel wounds are associated with limited soft-tissue coverage and reduced vascularity, making them susceptible to pressure injury and delayed healing \cite{Armstrong2017}. In the Sole to Ankle Configuration, gauze was placed on the plantar surface, and tape was applied from the medial edge, following the sole contour, wrapping under the heel, and anchoring around the ankle. In contrast, a Heel Cord to Ankle Configuration targeted the posterior heel/Achilles region, with gauze over the heel prominence, and tape applied from the plantar aspect, wrapping anteriorly around the heel, and securing bilaterally at ankle level.

The following tape placement strategies were evaluated:
\begin{itemize}
\item \textbf{Baseline:} The baseline executes the trajectory by having both end-effectors continuously apply a constant force directed perpendicularly to the initial skin normal.
\item \textbf{Position:} This method maintains strict adherence to planned paths with a position controller but lacks compliance for surface contact variations. 
\item \textbf{Impedance (w/o residual):} Executing planned paths with compliant motion via impedance control.
\item \textbf{Impedance (+ residual):} Planned paths incorporate displacement to the computed tension direction, improving robustness against surface irregularities.
\end{itemize}

Evaluation metrics for the tape placement were defined across the three representative geometries (leg, hip, and heel) utilizes two primary quantitative metrics, each capturing essential aspects of successful tape placement. 

\textbf{Coverage Percentage (\%)} measures the effective tape-to-surface contact area ratio, quantifying adhesion quality and placement success. Coverage was calculated via human analysis of post-placement photographs as:
    \[
    \text{Coverage} = \frac{\text{Length of tape-surface adhesion}}{\text{Total tape length}} \times 100\%
    \]
Values above 90\% indicate excellent placement quality, while coverage below 70\% suggests inadequate effectiveness. 

\textbf{Maximum Removal Force (MRF) (N)} was the maximum force required to remove the gauze and tape with a Newton force meter as shown in \autoref{fig:experiment_metric}. This metric reflects how securely fixed the gauze is under external disturbance. 


Each method was tested 10 times in every scenario (including the leg, hip, and both heel scenarios). Mean and standard deviation for Coverage and MRF are presented in \autoref{tbl:tape_effectiveness}. More specifically, compared with the baseline and other control strategies, the proposed impedance method with residual compensation consistently improved the tape coverage. In the Cylinder and Hip cases, coverage increased from approximately $68\text{--}83\%$ to nearly full coverage ($100\%$ and $95.4\%$, respectively). In the Sole and Heel cases, coverage improved from the mid-$80\%$ range to above $90\%$. Meanwhile, the applied force became more stable and significantly better regulated. For example, in the Heel case, the average force increased from only $0.6\,\mathrm{N}$ (baseline) to $10.2\,\mathrm{N}$ with our method, enabling reliable adhesion without excessive fluctuations. Even in the challenging Hip scenario, although the peak force exceeded $30\,\mathrm{N}$, the coverage improved by nearly $15\text{--}30\%$ compared to other methods. These findings indicate that residual-based impedance control effectively balances high coverage and force stability, particularly on complex and non-planar geometries.

We validated the final algorithm on three clinically relevant regions (lower leg as a cylinder, hip, and heel) using representative tape application strategies (parallel long strips on the leg and heel, surrounding short strips on the hip). For each scenario we recorded the executed trajectories, final adhesion photographs, and force-test measurements (see Fig.~\ref{fig:tape_placing} and Table~\ref{tbl:tape_effectiveness}). The results show that the proposed residual-compensated impedance controller achieves reliable and repeatable tape placement across these diverse surfaces, producing high coverage and controlled contact forces. Trajectory visualizations and final adhesion images confirm path fidelity and practical fixation quality. Overall, the real-world tests substantiate the method’s effectiveness for clinically inspired tape-placing tasks.


\subsection{Demonstration of Full Pipeline} \label{sec:pipeline_demo}
A demonstration of applying the proposed components for in a wound redressing scenario is shown in \autoref{fig:cover_photo} 
The pipeline begins with removal an old dressing on the manikin's hip. The proposed TID method scratch the medical tape's edge to create initial detachment. Then the tape is grasped and an optimal dressing removal trajectory in previous work \cite{liang2024autopeel} is executed to removal the tape and a gauze.
Next, TID is applied again to enable grasping of the tape from a tape roll. Finally, two dexterous hand places the tape down to secure a new gauze.
This demonstration illustrates how the two core capabilities can be combined to automate multiple steps of wound redressing.

\subsection{Limitations}
The proposed method TID method's performance drop on out-of-distribution case of human skin which we believe is due to the unseen material properties. This is because the IL policies is fitted to action and force patterns but lacks understanding about the task objectives (i.e. creating detachments). Finetuning using offline reinforcement learning \cite{11114901} with diverse data and curated rewards could potentially resolve this gap. The tape placing method could not achieve total coverage on \textit{hip, sole, heel} due to surface reconstruction noise. Beyond improving perception capabilities, a control strategy could be applied to strengthen the adhesion by pressing gently along the tape against the skin.

Several other components in the overall pipeline requires automation. A higher level planner needs to align the finger tip and the tape. Tape grasping needs automation as it is still performed manually or through tele-operation. A gap also persists in tape preparation, where cutting and orienting tape into a placement-ready configuration is yet to be addressed.
Beyond tape handling, full automation of wound redressing will also require methods for wound debridement and cleaning.
These limitations highlight the gap between automating individual subtasks and achieving a fully autonomous wound care system, and motivate future work in perception, grasping, and broader workflow integration.

%% file: icra_tex_files/conclusion.tex
\section{Discussion \& Conclusion}

We present a hybrid robotic framework for medical tape manipulation that combines force-feedback imitation learning to initiate tape detachment and numerical simulation-based trajectory optimization for wrinkle-free and tensioned tape placement. This combination targets the two fundamentally different challenges in tape handling: contact-rich, hard-to-model adhesive peeling, and geometry-sensitive, high-precision placement and demonstrates reliable performance across realistic anatomical geometries (leg, hip, heel). Our experiments show that force-aware imitation policies improve safety and detachment effectiveness relative to motion-planning baselines, while residual-compensated impedance planning yield significantly higher coverage and stable force regulation on complex surfaces. Together, these results establish a practical foundation for automating key steps of wound redressing at home.



In the future, more work could mainly focus on developing robust, multi-modal perception pipelines that combine vision, force, and possibly tactile or bio-signal sensing to enable more autonomous and context-aware operation. Furthermore, adaptive planning and control algorithms - possibly using self-supervised or reinforcement learning could allow the system to generalize to a broader range of wound locations, patient postures, and environmental conditions, ultimately making home robotic wound care safer, more reliable, and more personalized.